\documentclass[10pt,twocolumn,letterpaper]{article}

\usepackage{wacv}

\makeatletter
\@namedef{ver@everyshi.sty}{}
\makeatother
\usepackage{tikz}

\usepackage{times}
\usepackage{epsfig}
\usepackage{graphicx}
\usepackage{amsmath}
\usepackage{amssymb}
\usepackage{booktabs}
\usepackage{adjustbox}

\usepackage{comment}
\usepackage{amsmath,amssymb} 
\usepackage{color}
\usepackage{wrapfig}
\usepackage{graphicx}
\usepackage{float}

\wacvfinalcopy 

\ifwacvfinal
\usepackage[breaklinks=true,bookmarks=false]{hyperref}
\else
\usepackage[pagebackref=true,breaklinks=true,colorlinks,bookmarks=false]{hyperref}
\fi

\begin{document}

\title{LiveSeg: Unsupervised Multimodal Temporal Segmentation\\ of Long Livestream Videos}

\author{Jielin Qiu$^{1,2}$, Franck Dernoncourt$^{1}$, Trung Bui$^{1}$,
Zhaowen Wang$^{1}$, Ding Zhao$^{2}$, Hailin Jin$^{1}$ \\
$^{1}$Adobe Research, $^{2}$Carnegie Mellon University \\
  \small{$\left\{\text{jielinq,dingzhao}\right\}$@andrew.cmu.edu, $\left\{\text{dernonco,zhawang,bui,hljin}\right\}$@adobe.com}
}

\maketitle
\thispagestyle{empty}

\begin{abstract}
   Livestream videos have become a significant part of online learning, where design, digital marketing, creative painting, and other skills are taught by experienced experts in the sessions, making them valuable materials. However, Livestream tutorial videos are usually hours long, recorded, and uploaded to the Internet directly after the live sessions, making it hard for other people to catch up quickly. An outline will be a beneficial solution, which requires the video to be temporally segmented according to topics. In this work, we introduced a large Livestream video dataset named MultiLive, and formulated the temporal segmentation of the long Livestream videos (TSLLV) task. We propose LiveSeg, an unsupervised \textbf{Live}stream video temporal \textbf{Seg}mentation solution, which takes advantage of multimodal features from different domains. Our method achieved a $16.8\%$ F1-score performance improvement compared with the state-of-the-art method. 
\end{abstract}

\section{Introduction}
Video temporal segmentation has become increasingly important since it is the basis for many real-world applications, i.e., video scene detection, shot boundary detection, etc. 
Video temporal segmentation can be considered an essential pre-processing step, and an accurate temporal segmentation result could benefit many other tasks.
The video temporal segmentation methods lie in two directions: unimodal and multimodal approaches. Unimodal approaches only use the visual modality of the videos to learn scene change or transition in a supervised manner, while multimodal methods exploit available textual metadata and learn joint semantic representation in an unsupervised way.

\begin{figure}[htp]
  \centering
  \includegraphics[width=0.98\linewidth]{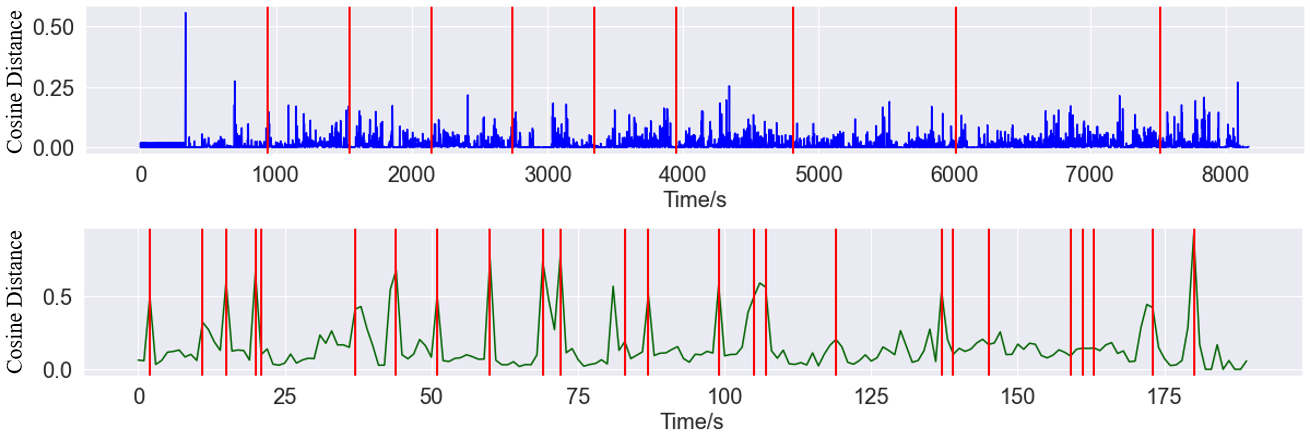}
  \caption{Comparison of temporal-pairwise cosine distance on visual features: (TOP) a Livestream video, (BOTTOM) a TVSum video (Blue \& Green: distance; Red: segment boundaries).}
  \label{Fig:cosdis_compare}
\end{figure}

A considerable amount of long Livestream videos are uploaded to the Internet every day, but it is challenging to understand the main content of the long video quickly. Traditionally, we can only have an inaccurate assumption by reading the video's title or using the control bar to manually access the video, which is time-consuming, inaccurate, and very easy to miss valuable information. An advantageous solution is to segment the long video into small segments based on the topics, making it easier for the users to navigate the content.

Most existing video temporal segmentation work focused on short videos. Some work explored movie clips extracted from long videos but easily segmented temporally by scene change. Jadon et al. proposed a summarization method based on the SumMe dataset \cite{Gygli2014CreatingSF}, which are 1-6 min short videos with clear visual change \cite{Jadon2020UnsupervisedVS}. When it comes to the long Livestream videos, the previous methods do not work well due to the extremely long length and new characteristics of the Livestream videos. So the critical problem is finding a practical approach to temporally segment the Livestream videos into segments. The quality of segmentation results can significantly impact further tasks. So here we propose a new task, TSLLV, temporal segmentation of long Livestream videos, which has not been explored yet. Different from other long videos, i.e., movies, Livestream videos usually contain more noisy visual information due to the visually abrupt change, and more noisy language information due to random chatting, conversational languages, and intermittent sentences, which means the content is neither clear nor well-organized, making it extremely hard to detect the segment boundaries. 
Comparison of the visual noisiness of the Livestream video and other videos and examples of Livestream transcripts are introduced in Section \ref{sec_dataset}.

To sum up, the main difficulties for temporally segmenting the Livestream videos are:\\
(1) The visual background remains similar for a considerable time, even though the topic has already changed, making the definition of boundaries ambiguous. For our MultiLive dataset collected from Behance\footnote{https://www.behance.net/live}, the hosts usually teach drawing or painting, where the main background is the board and remains similar for most parts of the video. Compared with movies, the movie's background changes dramatically when switching to another scene, so the Livestream videos can not be split directly based on visual scene change or transition differences. Fig.~\ref{Fig:cosdis_compare} shows an example comparison of temporal-pairwise cosine distance (distance between the $i$th frame and $i+1$th frame of the same video) on visual feature between a Livestream video and a TVSum video \cite{Song2015TVSumSW}, which shows the Livestream video's segment boundaries are not aligned with the visual scene change, making it difficult to segment.\\
(2) The visual change is neither consistent nor clear. As shown in Fig.~\ref{Fig:cosdis_compare}, there are abrupt changes in the visual site due to the host changing folders or zooming in/out, making the visual information extremely noisy. \\
(3) There is not enough labeled data for this kind of Livestream video, and it is challenging, time-consuming, and expensive to label them manually. Because it requires the human annotators to watch the entire video, understand the topics, and then temporally segment it, making it much more complicated than labeling images.

Our contributions are listed as follows:
\vspace{-5pt}
\begin{itemize}
    \item We introduced MultiLive, a new large dataset of Livestream videos, among which, 1,000 videos were manually segmented and annotated, providing human insights and references for evaluation.
    \vspace{-7pt}
    \item We formulate a new temporal segmentation of long Livestream videos (TSLLV) task according to the newly introduced MultiLive dataset.
    \vspace{-7pt}
    \item We proposed \textbf{LiveSeg}, an unsupervised \textbf{Live}stream temporal \textbf{Seg}mentation method by exploring multimodal visual and language information as a solution to TSLLV. We extract features from both modalities, explore the relationship and dependencies across domains, and generate accurate segmentation results. LiveSeg achieved a $16.8\%$ F1-score performance improvement compared with the SOTA method.
\end{itemize}

\section{Related Work}
\paragraph{Video Temporal Segmentation} Temporal segmentation aims at generating small segments based on the content or topics of the video, which is easy to achieve when the video is short or when the scene change is easy to detect, e.g., in movie clips. Previous works mainly focused on short videos or videos with clear scene changes, which is convenient to manually label a huge amount of videos as training sets for supervised learning \cite{Koprinska2001TemporalVS,Sidiropoulos2011TemporalVS,Zhou2013HierarchicalAC,Poleg2014TemporalSO,furnari2018personal,Sokeh2018SuperframesAT,Aakur2019APP}. 
\vspace{-10pt}
\paragraph{Action, Shot, and Scene Segmentation} Temporal action segmentation in videos has been widely explored \cite{Wang2019TemporalSN,Zhao2017TemporalAD,Lea2017TemporalCN,Gao2019VideoIS,Kuehne2020AHR,Sarfraz2021TemporallyWeightedHC,Wang2020BoundaryAwareCN}. 
However, those videos' characteristics are far different from Livestream videos, where the actions are well-defined, 
{the main goal is to group similar actions based on visual change, and the length of videos is much shorter,}
so the methods can not be adopted directly. Shot boundary detection task is also very relevant and has been explored in many previous works \cite{Hassanien2017LargescaleFA,Tang2018FastVS,Hato2019FastAF,Abdulhussain2018MethodsAC}, where shot is defined by the visual change. However, in Livestream videos, segments are not solely defined by visual information, the topics contained in language also contribute to the definition of each segment. Video scene detection is the most relevant task. However, previous methods only used visual information to detect the scene change \cite{Rao2020ALA,Rotman2016RobustAE,Rotman2018OptimallyGD,Chen2021ShotCS,Zhang2021BetterLS}, so the methods can not be adopted directly for Livestream videos either.
\vspace{-10pt}
\paragraph{Unsupervised Methods} Recently, unsupervised methods have also been explored for video temporal segmentation. \cite{Kanafani2021UnsupervisedVS} proposed incorporating multiple feature sources with chunk and stride fusion to segment the video, but the datasets used are still short videos \cite{Gygli2014CreatingSF,Song2015TVSumSW}. \cite{Fraser2020TemporalSO} used Livestream videos as materials. However, they used internal software usage as the segmentation reference, which is not available for most videos, making their method highly restricted. Because for most videos, we can only get access to visual and audio/language metadata. 
\vspace{-10pt}
\paragraph{Summary} Although previous models have shown reasonable results, they still suffer some drawbacks. Most work targeted short videos with clear scene changes instead of long videos, and only used visual information while ignoring other domains, like language. Due to the characteristics of the Livestream videos in our MultiLive dataset, methods that solely depend on visual features can not obtain accurate results, so a multimodal approach should be addressed to incorporate visual and language information.

\section{MultiLive Dataset}\label{sec_dataset}
\begin{figure}[htp]
  \centering
  \includegraphics[width=0.99\linewidth]{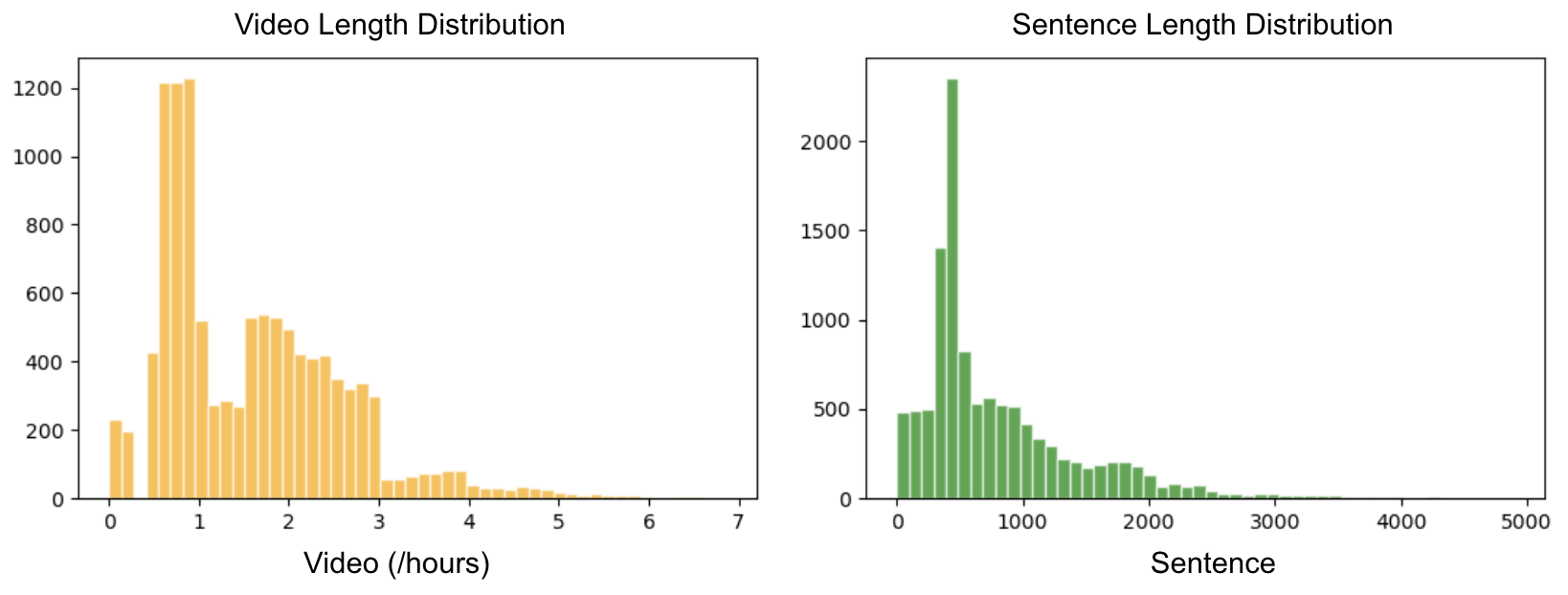}
  \caption{Histogram of MultiLive video length distribution and transcript length distribution (y-axis: number of videos).}
  \label{Fig:distribution}
\end{figure}

We introduced a large Livestream video dataset from Behance\footnote{https://www.behance.net/live} (the license is obtained and will be provided when the dataset is released), which contains Livestream videos for showcasing and discovering creative work. The dataset includes video ID, title, video metadata, extracted transcript metadata from audio signals (by Microsoft ASR \cite{Xiong2018TheM2}), offset (timestamp), duration of each sentence, etc. The whole dataset contains 11,285 Livestream videos with a total duration of 15,038.4 hours, the average duration per video is 1.3 hours. The entire transcript contains 8,001,901 sentences, and the average transcript length for each video is 709 sentences. (An example transcript is shown in the Appendix.)
The detailed statistics of the dataset are shown in Table~\ref{Statistics2} and Table~\ref{Statistics3}. 
From Tables \ref{Statistics2},\ref{Statistics3}, most videos are less than 3 hours and most videos' transcript contains less than 1,500 sentences.
In addition, we showed the  histogram of video length distribution and
transcript length distribution in Fig~\ref{Fig:distribution}.
\vspace{-8pt}
\begin{table}[htp]\small
    \centering
	\caption{Distribution of Livestream video duration.}
	\vspace{3pt}
	\begin{adjustbox}{width=0.7\linewidth}
		\begin{tabular}{c|c|c} 
			\hline
			Video Duration  &Number   &Percentage \\ \hline  
			0-1 h & 4,827 & 42.774\% \\ 
			1-2 h & 2,945 & 26.097\% \\ 
			2-3 h  & 2,523 & 22.357\% \\ 
			3-4 h & 705 & 6.247\% \\ 
			4-5 h & 210 & 1.861\% \\ 
			5-6 h & 70 & 0.620\% \\ 
			6-7 h & 11 & 0.097\% \\ \hline
	\end{tabular}
	\end{adjustbox}
	\label{Statistics2}
\end{table}
\vspace{-15pt}
\begin{table}[htp]\small
    \centering
	\caption{Distribution of transcript length.}
	\vspace{3pt}
	\begin{adjustbox}{width=0.75\linewidth}
		\begin{tabular}{c|c|c} 
			\hline
			Transcript Length  &~Number~   &~Percentage~ \\ \hline 
			0-500 & 5,512 & 48.844\% \\ 
			500-1,000 & 2,299 & 20.372\% \\ 
			1,000-1,500  & 1,890 & 16.748\% \\
			1,500-2,000 & 989 & 8.746\% \\ 
			2,000-2,500  & 365 & 3.234\% \\
			2,500-3,000 & 118 & 1.046\% \\ 
			3,000-3,500 & 84 & 0.744\% \\ 
			3,500-4,000 & 35 & 0.310\% \\ 
			4,000-4,500 & 12 & 0.106\% \\ 
			4,500-5,000 & 3 & 0.027\% \\ \hline
	\end{tabular}
	\end{adjustbox}
	\label{Statistics3}
\end{table}
\vspace{-15pt}

Besides, for the purpose of evaluation, we provide human annotations of 1,000 videos with segmentation boundaries annotated manually by human annotators for evaluation. The human annotators are asked to watch and understand the whole video and split each into several segments based on their understanding of the video content. 
{The current 1,000 videos' annotation includes 10 annotators from Amazon Mechanical Turk \footnote{https://www.mturk.com/} (legal agreement signed). The annotators were separated into groups and each group watched part of the videos and then discussed the results together about the segmentation results to ensure the quality of the annotation was agreed upon by all the annotators. They were instructed to pay more attention to topic change, w.r.t. the moment that the live-streamer starts discussing a different topic. } 

\begin{table}[htp]\small
    \centering
	\caption{Comparison of MultiLive with existing datasets.}
	\vspace{3pt}
	\begin{adjustbox}{width=0.99\linewidth}
		\begin{tabular}{c|cccc}  
			\hline
			\textbf{Statistics}  &MultiLive &SumMe \cite{Gygli2014CreatingSF} & TVSum \cite{Song2015TVSumSW} &OVP \cite{Avila2011VSUMMAM}     \\  \hline
			Labeled videos & 1,000  &25  &50  & 50    \\ \hline
			Ave. length (min) & 78 mins  & 2.4 mins & 4.2 mins & 1.5 mins   \\ \hline
			Ave. scene num  &8.8  &5.1  &52.2   &8.8    \\ \hline
			Ave. SLR \\(min/scene)  &8.86  & 0.47  &0.08   &0.17   \\ \hline
			Ave. SD   &0.07  &0.22   &0.19   &0.35   \\ \hline
	\end{tabular}
	\end{adjustbox}
	\label{table:compare_dataset}
\end{table}

There are several widely used video datasets in temporal segmentation or video summarization tasks \cite{Gygli2014CreatingSF,Song2015TVSumSW,Avila2011VSUMMAM}, Table~\ref{table:compare_dataset} shows the comparison of our dataset with the others. The amount of labeled videos of the others is less than 50, while we provide human annotations for 1,000 videos. The average length of the videos from our dataset is much longer than others, while the number of segments is in the same order of magnitude or even smaller than the others. The effect is that, the average SLR (scene length ratio) of the Livestream dataset is much larger, where average SLR (scene length ratio) can be considered a metric to represent the average length of each scene in the video, calculated by (ave.length / ave. scene num). So the larger the ratio, the more content contained in each segment, leading to more difficulty finding the segment boundaries.

\begin{figure}[htp]
  \centering
  \includegraphics[width=0.99\linewidth]{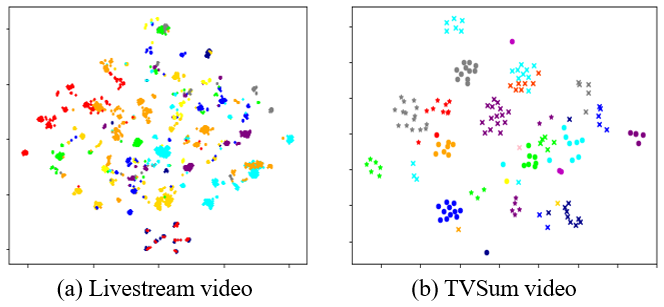}
  \caption{(a) Visual features of a Livestream video; (b)Visual features of a TVSum video, where different colors represent different segments within one video.}
  \label{Fig:visual-compare}
\end{figure}
\begin{figure}[H]
  \centering
  \includegraphics[width=0.99\linewidth]{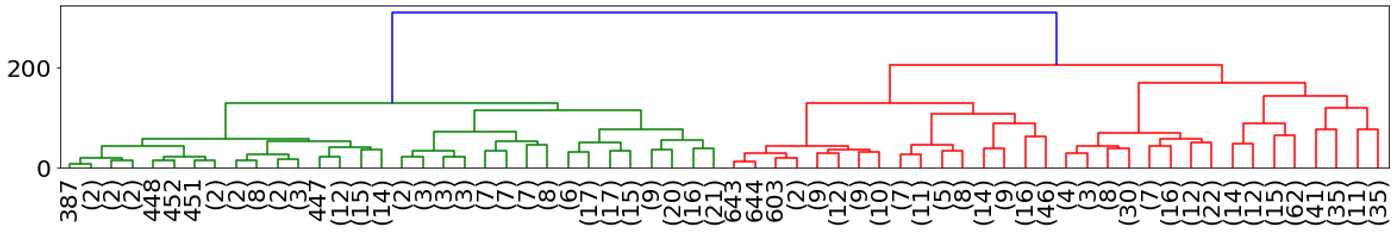}
  \caption{Dendrogram result of one Livestream video by hierarchical clustering of visual features, where the numbers below the bottom layer represent the number of images belongs to the corresponding sub-tree.}
  \label{Fig:dendrogram_result}
\end{figure}

To demonstrate a more precise understanding of the visual information of Livestream videos, we compared the visual features extracted from one example Livestream video and one example TVSum video \cite{Song2015TVSumSW}. We extracted video frames from the raw video sequence, used ResNet50 model \cite{He2016DeepRL} (pre-trained on ImageNet) to extract the visual features of each video frame, and adopted t-SNE \cite{Maaten2008VisualizingDU} to visualize the visual features. 
Fig. \ref{Fig:visual-compare}(a) shows the Livestream video's visual feature distribution, different colors with the same marker ``o" representing different segments, ten segments in total. We can find that feature points which belong to different segments mix together and thus are hard to separate. As for TVSum's video result in Fig. \ref{Fig:visual-compare}(b), different color or different marker ``o"/``x"/``$*$" all represents different segments, 23 segments in total, which shows the points belong to different segments can be distinguished more easily than the Livestream video. This proves our statement that Livestream videos contain more noisy visual information, making it much harder to be temporally segmented by traditional methods. 
\vspace{-5pt}
\begin{table}[H]\small
    \centering
	\caption{Comparison of different type of videos.}
		\begin{tabular}{c|c|c}  
			\hline
			\textbf{Statistics}  &ASR WER & USR  \\ \hline 
			Film Corpus  \cite{Walker2011PerceivedON,Walker2012AnAC} &0.01   &0.126   \\
			Movie Dialog Corpus \cite{MovieDialog}  &0.01   &0.139  \\
			MultiLive  &0.05   &0.458    
			\\ \hline
	\end{tabular}
	\label{table:compare_type}
\end{table}
\vspace{-10pt}
Table~\ref{table:compare_type} also shows the comparison of our Livestream data with movie datasets \cite{Walker2011PerceivedON,Walker2012AnAC,MovieDialog}, which were collected from IMDB and TMDB, to emphasize the differences between Livestream videos and movies. 
Table~\ref{table:compare_type} shows the Livestream videos' ASR WER (word error rate) is higher than movies, and the USR (unrelated sentence rate) is much higher than movies, which contain more meaningless conversational languages.
We further used hierarchical clustering to group the frames based on visual features and generated a dendrogram. As shown in Fig.~\ref{Fig:dendrogram_result}, we could find that the video frames far away from each other in timestamp can still be clustered together into the same group if only visual features are used. It supports the claim that using only visual information is insufficient to generate accurate temporal segmentation results, as the visual domain lacks sufficient information. So other domain features should be explored to provide more information.

\vspace{-10pt}
{To show a representative comparison, we computed the frame-level average distance (ave. SD) between the segments of our MultiLive dataset and the SumMe, TVSum, and OVP datasets. The results are shown in Table~\ref{table:compare_dataset}. The distance is computed on the two adjacent frames on each video segment boundary (last frame of $i$th segment, and first frame of $(i+1)$th segment, and the average results could show the average visual difference comparison. As in Table~\ref{table:compare_dataset}, we can find that the ave. SD of the MultiLive dataset is much smaller than the ave. SD of other datasets, which could be a representative metric to demonstrate that Livestream video's visual change is much more noisy than existing datasets, making it more difficult to segment.}

\section{LiveSeg: Unsupervised Multimodal Temporal Segmentation of Livestream Videos}
\begin{figure*}[htp]
  \centering
  \includegraphics[width=0.86\linewidth]{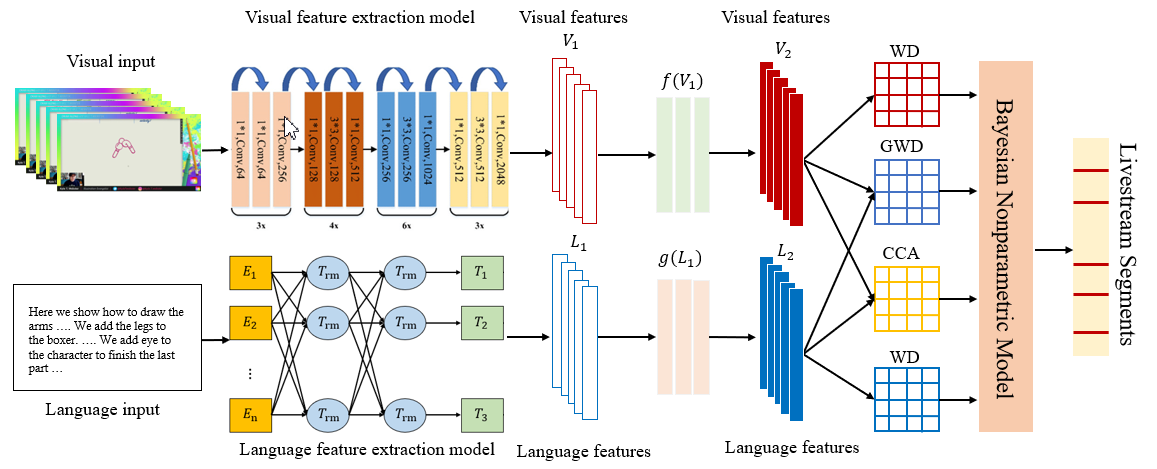}
  \caption{The framework of LiveSeg for unsupervised multimodal Livestream video temporal segmentation.}
  \label{Fig:seg_model}
\end{figure*}

The TSLLV task (temporal segmentation of long Livestream video) aims to accurately and temporally segment the Livestream videos based on the topics. Due to the absence of segmented labels and the time-consuming of manually labeling a huge amount of such long videos, we adopt unsupervised methods to segment the Livestream videos temporally. The whole framework is shown in Fig.~\ref{Fig:seg_model}. 
{Given a Livestream video ${\cal S}$, our target is to temporally segment video $\cal S$ into [${S}_1, S_2, ... , S_{k}$] based on topics, where $k$ is the number of segments. The only available materials are video (visual input) and transcripts (language input). The number of segments of each to-be-segmented video is not preliminary given.}

\subsection{LiveSeg Framework}\label{segmentation_sec}
The LiveSeg model takes input from the visual domain and the language domain. For visual features, we sample video frames [$f_1, f_2, ... , f_n$], where $n$ is the timestamp, from the raw video $\cal S$ (one frame per second to reduce the computation complexity). Then we use ResNet-50 \cite{He2016DeepRL} pre-trained on ImageNet \cite{Russakovsky2015ImageNetLS} to extract visual features (fingerprints) $V_1 = [V_{11}, V_{12}, ... , V_{1n}]$, where the visual fingerprints represent the video content. For the language features, due to the fact that the transcript is not temporally perfectly aligned with video frames, we first assign the transcript sentences to the corresponding video frame. If there are overlaps between several sentences or several frames, we duplicate those in a corresponding manner, and formulate frame-transcript pairs for each sampled frame in the timeline. 
Since the frames are sampled by a one-second time window, the transcripts are also aligned with each time window. If one transcript sentence $T_i$ does not end for the given window, meaning the language has overlapped with two adjunct time windows, then we will assign this sentence $T_i$ to both time window $t$ and time window $t+1$.
We then extract sentence embeddings with BERT \cite{Devlin2019BERTPO} to get sentence-level representations $L_1 = [L_{11}, L_{12}, ... , L_{1n}]$. The embedding model used in our formulation is ``all-MiniLM-L6-v2" from Sentence-Transformers  \cite{reimers-2019-sentence-bert}, which is trained on large sentence level datasets using a self-supervised contrastive learning objective from pre-trained model \cite{Wang2020MiniLMDS} and fine-tuned on a sentence pairs dataset. Due to the ambiguity of the transcript, i.e., the examples are shown in the Appendix, redundant and noisy words are removed before generating language embeddings (redundant and noisy words mean the words that appear more than three times in a row due to the live-streamer's speaking error). 

The previous work \cite{Liu2019AligningVR,Chen2020GraphOT,Jia2021ScalingUV} which took advantage of the alignment of vision and language features inspired us to assume that there should be a relationship and dependency between visual and language features.
{\cite{Chen2019ImprovingSL,Lin2021MakingTM,Wang2021UFOAU,Chen2020UNITERUI,Kim2021ViLTVT} find that Optimal Transport shows tremendous power in sequence-to-sequence learning. In addition, \cite{Chen2020GraphOT,Yuan2020AdvancingWS} find that Gromov Wasserstein Distance shows even better performance in measuring the distances in counterpart domains. Canonical Correlation Analysis, a well-known approach to exploring the correlation between different modalities, has been studied in many previous works for its ability to recognize the cross-domain relationship \cite{Andrew2013DeepCC,Yan2015DeepCF,Xu2015JointlyMD,Guo2019CanonicalCA}.
\cite{Teh2008HierarchicalBN,Nagano2019HVGHUS} showed that Bayesian Nonparametric Models performed well on temporal segmentation task, especially under unsupervised settings, which stands as a good candidate for our TSLLV task.
}
Therefore, we adopt Deep Canonical Correlation Analysis \cite{Andrew2013DeepCC} to encode the dependency for a hierarchical feature transformation. The networks transform raw visual features $V_1$ to high-level visual features $V_2$ with the transformation $f(V_1)$, and transform raw language features $L_1$ to high-level language features $L_2$ with the transformation $g(L_1)$. 
Then we compute the Wasserstein Distance (WD) on the high-level temporal visual features $V_2$ and language features $L_2$. We also calculate the Gromov Wasserstein Distance (GWD) and Canonical Correlation Analysis (CCA) on the two different modalities at the same timestamp, then use Bayesian Nonparametric models \cite{Johnson2013BayesianNH} to segment the Livestream videos temporally. The details of each part are introduced in
the following paragraphs and sections. More details about WD, GWD, and CCA can also be found in the Appendix.
\vspace{-10pt}
\paragraph{Wasserstein Distance}\label{sec:WD}
Wasserstein Distance (WD) is introduced in Optimal Transport (OT), which is a natural type of divergence for registration problems as it accounts for the underlying geometry of the space, and has been used for multimodal data matching and alignment tasks \cite{Chen2020GraphOT,Yuan2020AdvancingWS,Lee2019HierarchicalOT,Demetci2020GromovWassersteinOT,Han2022AnEE,Qiu2022SemanticsConsistentCS}. In Euclidean settings, OT
introduces WD $\mathcal{W}(\mu, \nu)$, which measures the minimum effort required to “displace” points across measures $\mu$ and $\nu$, where $\mu$ and $\nu$ are values observed in the empirical distribution. In our setting, we compute the temporal-pairwise Wasserstein Distance on both visual features and language features, considering each feature vector representing each frame or transcript embedding.
The temporal-pairwise WD on both visual and language features encodes the temporal difference and consistency within the same domain.
\vspace{-10pt}
\paragraph{Gromov Wasserstein Distance}\label{sec:GWD}
Classic OT requires defining a cost function across domains, which can be challenging to implement when the domains are in different dimensions \cite{Redko2020COOptimalT}. Gromov Wasserstein Distance (GWD) \cite{Peyr2016GromovWassersteinAO} extends OT by comparing distances between samples rather than directly comparing the samples themselves. 
In our framework, the computed GWD across domains is to capture the relationship and dependencies between visual and language domains.
\vspace{-10pt}
\paragraph{CCA and DCCA}\label{sec:CCA}
Canonical Correlation Analysis (CCA) is a method for exploring the relationships between two multivariate sets of variables, which can learn the linear transformation of two vectors in order to maximize the correlation between them, which is used in many multimodal problems \cite{Andrew2013DeepCC,Qiu2018MultiviewER,Liu2019MultimodalER,Bao2019InvestigatingSD,Gong2013AME,Liu2022ComparingRP}. In our problem, we apply CCA to capture the cross-domain relationship of visual features $V_{2l}$ and language features $L_{2l}$.
To obtain $V_2$ and $L_2$, DCCA is applied in the framework for nonlinear feature transformation. 
The parameters are trained to optimize this quantity using gradient-based optimization by taking the correlation as the negative loss with backpropagation to update the nonlinear transformation model \cite{Andrew2013DeepCC}. More details can be found in the Appendix.

\subsubsection{Bayesian Nonparametric Model}\label{sec:HSMM}
We used Hierarchical Dirichlet Process Hidden semi-Markov Model (HDP-HSMM) to generate the video segments for modeling \cite{Johnson2013BayesianNH,Fried2020LearningTS}, which can infer arbitrarily large state complexity from sequential and time-series data. More discussion about HMM, HSMM, and their drawbacks are introduced in the Appendix.  

\begin{figure}[htp]
\centering
\includegraphics[width=0.43\textwidth]{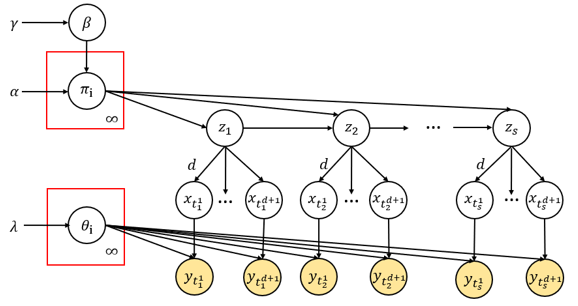}
\caption{\footnotesize Graphical model of HDP-HSMM}
\label{Fig:HSMM}
\end{figure}

The process of HDP-HSMM is illustrated in Fig.~\ref{Fig:HSMM}. In the model, $z_i$ denotes the classes of the segments, $\beta$ denotes an infinite-dimensional multinomial distribution, which is generated from the GEM distribution and parameterized by $\gamma$ \cite{Pitman2002PoissonDirichletAG}. GEM denotes the co-authors Griffiths, Engen, and McCloskey, with the so-called stick-breaking process (SBP) \cite{Sethuraman1991ACD}. The probability $\pi_i$ denotes the transition probability, which is generated by the Dirichlet process and parameterized by $\beta$ \cite{Teh2006HierarchicalDP}:
\begin{equation}\footnotesize
\beta \mid \gamma \sim \operatorname{GEM}(\gamma),~~
\pi_{i} \mid \alpha, \beta \sim \operatorname{DP}(\alpha, \beta), i=1,2, \cdots, \infty 
\end{equation}
where $\gamma$ and $\alpha$ are the concentration parameters of the Dirichlet processes (DP). The probability distribution is constructed through a two-phase DP named Hierarchical Dirichlet process (HDP) \cite{Teh2006HierarchicalDP,Nagano2019HVGHUS}. The class $z_i$ of the $i$th segment is determined by the class of the $(i-1)$th segment and transition probability $\pi_i$. 

In HSMM, state transition probability from state $i$ to $j$ can be defined as $\pi_{i, j}=p\left(x_{t+1}=j \mid x_{t}=i\right)$, then the transition matrix can be denoted as $\pi=\left\{\pi_{i, j}\right\}_{i, j=1}^{|\chi|}$, where $|\chi|$ denotes the number of hidden states. The distribution of observations $y_t$ given specific hidden states is denoted by $p\left(y_{t} \mid x_{t}, \theta_{i}\right)$, where $\theta_i$ denotes the emission parameter of state $i$. Then the HSMM can be described as:
\vspace{-10pt}
\begin{equation}
\begin{aligned}
x_{s} \mid x_{s-1} \sim \pi_{x_{s-1}}, d_{s} \sim g\left(\omega_{s}\right),
y_{t} \mid x_{s}, d_{s} \sim F\left(\theta_{x_{s}}, d_{s}\right)\\
\end{aligned}
\end{equation}
where $F(\cdot)$ is an indexed family of distributions, the probability mass function of $d_s$ is $p\left(d_{t} \mid x_{t}=i\right)$, $g\left(\omega_{s}\right)$ denotes a state-specific distribution over the duration $d_s$, and $\omega_s$ denotes the parameter priori of the duration distributions.

In HDP, let $\Theta$ be a measurable space with a probability measure $H$ on the space, $\gamma$ is a positive real number named the concentration parameter. $\operatorname{DP}(\gamma, H)$  is defined as the distribution of the random probability measure of $G$ over $\Theta$. For any finite measurable partition of $\Theta$, the vector  is distributed as a finite-dimensional Dirichlet distribution:
\begin{equation}\footnotesize
G_{0}\sim \operatorname{DP}(\gamma, H), G_{0}=\sum_{k=1}^{K} \beta_{k} \delta_{\theta_{k}},\theta_{k} \sim H, \beta \sim \operatorname{GEM}(\gamma)
\end{equation}
where $\theta_k$ is the distribution of $H$, $\beta \sim \operatorname{GEM}(\gamma)$ represents the stick-breaking construction process of the
weight coefficient \cite{Paisley2010ASC,Zhou2011HierarchicalDP}, and $\delta_{\theta}$ is the Dirac function. The model can be written as:
\begin{equation}\footnotesize
\theta_{i} \sim H(\lambda), i=1,2, \cdots, \infty, ~~
z_{s} \sim \bar{\pi}_{z_{s-1}}, s=1,2, \cdots, S 
\end{equation}
\begin{equation}\footnotesize
D_{s} \sim g\left(\omega_{z_{s}}\right), s=1,2, \cdots, S, ~~
\omega_{i} \sim \Omega
\end{equation}
\begin{equation}\small
x_{t_{s}^{1}: t_{s}^{D_{s}+1}}=z_{s}, ~~y_{t_{s}^{1}: t_{s}^{D_{s}+1}} \sim \mathrm{F}\left(\theta_{x_{t}}\right)
\end{equation}
where $\pi_i$ is the distribution parameter of hidden state sequence $z_s$, implying that HDP provides an infinite number of states for HSMM, $D_s$ is the length distribution of the state sequence with distribution parameter $\omega$, and $y_{t_s}$ is the observation sequence with distribution parameter $\theta_i$ \cite{Qiao2021CharacterizationOT}.

For parameter inference of the HDP-HSMM model, a  weak-limit Gibbs sampling algorithm is applied \cite{Johnson2013BayesianNH}. The weak limit approximation transforms the infinite dimension hidden state into finite dimension form so that the hidden state chain can be updated according to the observation data \cite{Qiao2021CharacterizationOT}. It is assumed that the basic distribution $H(\cdot)$ and the observation series distribution $F(\cdot)$ are conjugated distributions, the hidden states distribution $g(\cdot)$ is a Poisson distribution, and the hidden states distribution and the observation series distribution are independent. We first sample the weight coefficient $\beta$ and the state sequence distribution parameter $\pi_i$:
\begin{equation}\footnotesize
\beta \mid \gamma \sim \operatorname{Dir}(\gamma / S, \cdots, \gamma / S)
\end{equation}
\begin{equation}\footnotesize
\pi_{i} \mid \alpha, \beta \sim \operatorname{Dir}\left(\alpha \beta_{1}, \cdots, \alpha \beta_{s}\right) j=1, \cdots S
\end{equation}
Then we sample the observation distribution parameters $\theta_i$ and state duration distribution parameter $\omega_i$ according to observation data. It is assumed that the observed data obey a multivariate Gaussian distribution, the model parameters $\theta_{i}=\left(u_{i}, \Sigma_{i}\right)$ obey the Normal–Inverse–Wishart distribution:
\begin{equation}\footnotesize
\operatorname{NIW}\left(u, \Sigma \mid v_{0}, \Delta_{0}, \mu_{0}, S_{0}\right) \triangleq \mathrm{N}\left(\mu \mid \mu_{0}, S_{0}\right) * \operatorname{IW}\left(\Sigma \mid v_{0}, \Delta_{0}\right)
\end{equation}
where $\varphi=\left\{u_{0}, S_{0}, v_{0}, \Delta_{0}\right\}$ are prior parameters, $\mu_0$ and $S_0$ are the prior mean and co-variance matrices, and $\nu_0$ and $\Delta_0$ are the degrees of freedom and scale of NIW distribution. The state duration distribution is a Poisson distribution, and parameter $\omega_i$ follows a Beta distribution: 
$\omega \sim \operatorname{Beta}\left(\eta_{0}, \sigma_{0}\right)$.
Then we update parameters according to the observation data \cite{Johnson2013BayesianNH,Fox2009BayesianNL}.

\subsubsection{Final Segmentation Boundaries}\label{sec:post}
The raw output from the Bayesian nonparametric model contains both short and long segments, but the short segment may not contain comprehensive information, which will be useless as the final results. 
{So we used a heuristic-based method to group the small segments into the large ones. The method is straightforward,  where a parameter $l_s$ defines the minimum length of the generated segments, if there is a segment shorter than $l_s$, then we compute the visual and textual similarity of this small segment with the two adjacent segments, and group the small segment into the one with higher similarity. Since these small segments are mostly due to the live-streamer abruptly zooming in/out or randomly chatting about something unrelated to the main topic, which has little influence on the segmentation results (i.e., a small part inside a big chunk), we just used this simple method to make the results look more cleaner.  } 
The method introduced a parameter $l_s$, defined as the minimum length of the generated segments, which is used to hierarchically group the generated small segments into the bigger ones to eliminate the effect caused by small segments.

\subsection{Baseline Methods}\label{sec_baseline}
We select several strong and representative baseline methods for comparison, which include:
\vspace{-5pt}
\begin{itemize}
    \item \textbf{Hierarchical Cluster Analysis (HCA)} HCA aims at finding discrete groups with varying degrees of similarity represented by a similarity matrix \cite{Everitt2010ComprarTC,CohenAddad2018HierarchicalCO}, which produces a dendrogram as the intermediate result. The distance for the Livestream video setting is defined as: $d = \alpha_b d_{t} + (1- \alpha_b) d_{f}$, where $d_{t}$ is the timestamp distance, $d_{f}$ is the feature content distance, and $\alpha_b$ is used to balance the feature distance and time distance. Feature points representing content get separated further apart when the time distance of corresponding features is large. Another parameter $\beta_b$ controls the number of generated segments. $\beta_b$ is compared with the similarity score $\tt sim$ to cut through the dendrogram tree at a certain height (w.r.t $\beta_b$ = $\tt sim$) to create clusters of images.
    \vspace{-5pt}
     \item \textbf{TransNet V2}  Soucek et al. proposed TransNet V2 model for shot transition detection \cite{Souvcek2020TransNetVA}, which can also generate segmentation results and showed better performance than previous method \cite{Tang2018FastVS}.
     \vspace{-5pt}
    \item \textbf{Hecate} Song et al. proposed the Hecate model to generate thumbnails, animated GIFs, and video summaries from videos \cite{Song2016ToCO}, where shot boundary detection is one of the steps. This step will be used to compare with the other baseline methods as well as our method.
    \vspace{-5pt}
     \item \textbf{Optimal Sequential Grouping (OSG)} Rotman et al. proposed video scene detection algorithms based on the optimal sequential grouping \cite{Rotman2016RobustAE,Rotman2018OptimallyGD}, which included finding pairwise distances between feature vectors and splitting shots into non-intersecting groups by optimizing a distance-based cost function.
     \vspace{-5pt}
     \item \textbf{LGSS} Rao et al. proposed a local-to-global scene segmentation framework (LGSS) \cite{Rao2020ALA}, which used multiple features extracted by ResNet50, Faster-RCNN \cite{Ren2015FasterRT}, TSN \cite{Wang2016TemporalSN}, and NaverNet \cite{Chung2019NaverAA}. The temporal segmentation step is based on PySceneDetect \cite{PySceneDetect}. 
\end{itemize}

\section{Experiments and Results}

\subsection{Temporal Segmentation on MultiLive Dataset}\label{sec_result_seg}

For Livestream videos, the raw visual feature dimension is 2,048, and the raw language feature dimension is 384 extracted by pre-trained-BERT models from  Huggingface\footnote{https://huggingface.co/}. For the hierarchical transformation performed by DCCA, the network architecture and parameters are shown in the Appendix. In our experiments, we set $l_s$ to one minute. {To make a fair comparison, we also applied the post-processing step in Sec~\ref{sec:post} to all the baselines.}

\begin{figure}[htp]
  \centering
  \includegraphics[width=0.99\linewidth]{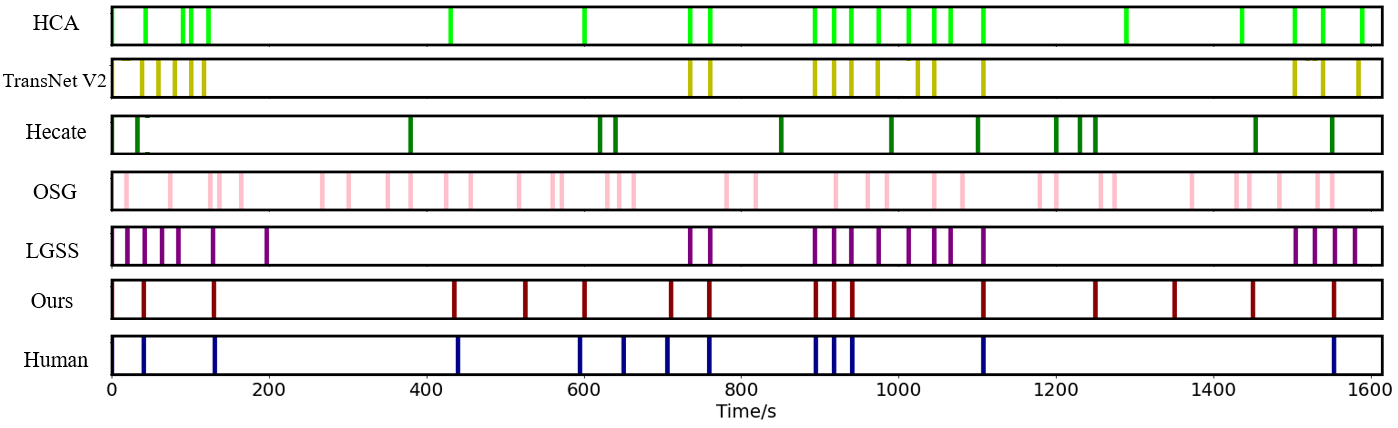}
  \caption{Comparison of boundary candidates by different methods, from top to bottom: (1) HCA \cite{CohenAddad2018HierarchicalCO}, (2) TransNet V2 \cite{Tang2018FastVS}, (3) Hecate \cite{Song2016ToCO}, (4) OSG \cite{Rotman2018OptimallyGD}, (5) LGSS \cite{Rao2020ALA}, (6) ours (LiveSeg), and (7) Human Annotations.}
  \label{Fig:compare}
\end{figure}

Due to the characteristic of Livestream videos, the video frames and transcripts are not perfectly aligned with the segments. Besides, different people segment the same video differently due to human preferences, which also needs to be considered. We performed the TSLLV task using baseline methods in Section \ref{sec_baseline} and our LiveSeg method on the MultiLive dataset. We evaluate the performance of different methods on the 1,000 annotated videos. Comparison results of baseline methods, our method, and human annotations for one Livestream video are shown in Fig.~\ref{Fig:compare}. We can see that scene transition detection methods will generate inaccurate segments as the visual change is noisy for Livestream videos. However, many essential boundaries will be missed if simply improving the clustering threshold. Compared with existing methods, our results are more accurate and can be comparable with human annotations. 

For quantitative analysis, tolerance interval $\omega_t$ is introduced. The correctness of the segmentation is judged at each position of this interval: a false alarm is declared if the algorithm claims a boundary in the interval while no reference boundary exists in the interval, and a miss is declared if the algorithm does not claim a boundary in the interval while a reference boundary exists in the interval \cite{Fiscus1999NISTs1T}. In our experiment, we set $\omega_t$ to one minute, and we adopt precision, recall, and F1-score  metrics to compare the performance of our results with human annotations. 
As shown in Table~\ref{Table:seg_results}, our segmentation results outperform other baseline results. Besides, considering modality, multimodal segmentation outperforms single modality results, showing that the relationship learned between the visual domain and the language domain can truly benefit temporal segmentation.
\vspace{-5pt}
\begin{table}[htp]\small
\centering
	\caption{Comparison of segmentation results.}
	\begin{adjustbox}{width=0.99\linewidth}
		\begin{tabular}{cccccc} 
			\hline
			Methods  &Backbone &Modality &Precision   &Recall &F1-score\\ \hline   
			HCA \cite{CohenAddad2018HierarchicalCO} &HCA  &Visual  &0.482  &0.487  &0.484\\ 
			TransNet V2 \cite{Tang2018FastVS} &ResNet-18  &Visual  &0.536   &0.525  &0.530  \\ 
			Hecate \cite{Song2016ToCO} &Clustering  &Visual  &0.539  &0.533  &0.536  \\ 
			OSG \cite{Rotman2018OptimallyGD} &DP  &Visual  &0.574  &0.557 &0.565   \\ 
			LGSS \cite{Rao2020ALA} &Bi-LSTM  &Visual   &0.587   &0.581  &0.584   \\ \hline
			LiveSeg-Visual &LiveSeg  &Visual  &0.591   &0.666  &0.626   \\ 
			LiveSeg-Language &LiveSeg  &Language  &0.589   &0.568  &0.578   \\ 
			LiveSeg-Multimodal &LiveSeg  &Multimodal  & \textbf{0.673}  & \textbf{0.697} & \textbf{0.685}   \\ \hline
	\end{tabular}
	\end{adjustbox}
	\label{Table:seg_results}
\end{table}
\vspace{-10pt}
\subsection{Ablation Study}
Multiple heuristics could have an impact on the segmentation performance, such as tolerance interval $\omega_t$ and the parameters in the Bayesian nonparametric model. We carried out several ablation experiments on the influence of different parameters with multimodal features, where the results are shown in  Table~\ref{Table:window} and Fig.~\ref{Fig:ablation}. More detailed results are shown in the Appendix due to the page limit.
\vspace{-5pt}
\begin{figure}[htp]
  \centering
  \includegraphics[width=0.99\linewidth]{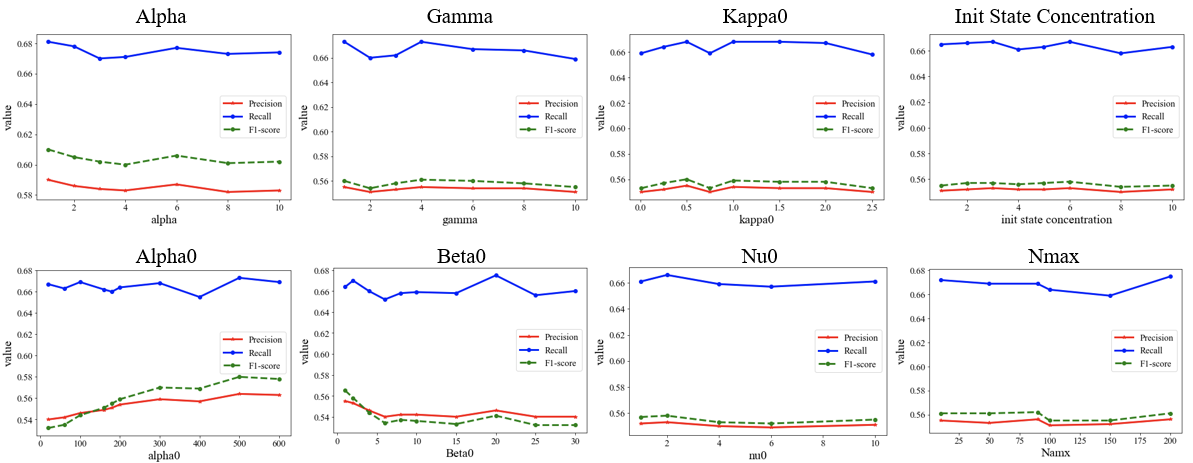}
  \caption{Segmentation performance with different parameters (Red: precision; Blue: recall; Green: F1-score).}
  \label{Fig:ablation}
\end{figure}
\vspace{-15pt}
\begin{table}[htp]\small
    \centering
	\caption{Comparison of performance with different interval $\omega_t$.}
	\vspace{5pt}
		\begin{adjustbox}{width=0.6\linewidth}
		\begin{tabular}{cccc}  
			\hline
			$\omega_t$ &Precision   &Recall &F1-score\\ \hline  
			0.5 min  & 0.608 &0.672  &0.627 \\ 
			1.0 min  & 0.673  &0.697  & 0.685\\ 
			1.5 min  & 0.605 & 0.666  & 0.621 \\ 
			2.0 min  & 0.600 & 0.659 & 0.615 \\ 
			2.5 min  & 0.598 & 0.653 & 0.610 \\ 
			3.0 min  & 0.595 & 0.647 & 0.605 \\ \hline
	\end{tabular}
	\end{adjustbox}
	\label{Table:window}
\end{table}

We also provided an ablation study on different components, since only using WD is the same as LiveSeg-Visual and LiveSeg-Language, so we provide additional ablation study results on GWD and CCA. In Table~\ref{Table:ablation_GWD}, we can find that combining all of them (LiveSeg-Multimodal) can achieve better performance than using only one of the components.

\begin{table}[htp]\small
    \centering
    \caption{Ablation study of different components.}
    	\begin{adjustbox}{width=0.85\linewidth}
		\begin{tabular}{c|c|c|c} 
			\hline
			  &Precision &Recall  &F1-score    \\ \hline 
			GWD  &0.622  &0.673   & 0.646    \\ \hline
			CCA  &0.603  &0.654   & 0.615     \\ \hline
			WD (LiveSeg-Visual)  &0.591  &0.666   & 0.605    \\ \hline
			WD (LiveSeg-Language)  &0.589  &0.568  & 0.606     \\ \hline
	\end{tabular}
	\end{adjustbox}
	\label{Table:ablation_GWD}
\end{table}

\subsection{Comparison on Other Datasets}
In addition, we compare our method with state-of-the-art unsupervised video summarization method \cite{Apostolidis2021ACSUMGANCA} on the famous video summarization benchmark datasets,  SumMe \cite{Gygli2014CreatingSF} and TVSum \cite{Song2015TVSumSW}. 
{We used the same key-fragment-based approach for evaluation \cite{Zhang2016VideoSW}, where the similarity between a machine-generated and a user-defined ground-truth summary is represented by expressing their overlap using the F-Score. For a given video and a machine-generated summary, this protocol matches the latter against all the available user summaries for this video and computes a set of F-Scores. More details of this experiment are shown in the Appendix. }
Table~\ref{Table:compare_summe} shows the comparison F1-score of our method with SUM-GAN \cite{Apostolidis2021ACSUMGANCA}, our method can still show slightly better results on the SumMe dataset and competitive results on the TVSum dataset, which clearly demonstrated the effectiveness of our method.

\begin{table}[htp]\small
    \centering
	\caption{Comparison with SOTA unsupervised baseline on traditional video summarization datasets.}
	\vspace{5pt}
    \centering
		\begin{tabular}{c|c|c} 
			\hline
			F1-score  &LiveSeg &SUM-GAN \cite{Apostolidis2021ACSUMGANCA}     \\ \hline 
			SumMe  &51.3  &50.8      \\ \hline
			TVSum  &60.9  &60.6       \\ \hline
	\end{tabular}
	\label{Table:compare_summe}
\end{table}
\vspace{-10pt}
\section{Discussion of Limitation and Future Work}
The current method targets long Livestream videos, which shows better performance than existing ones, given that the current setting of both visual input and language input are highly noisy. However, it may not work better than supervised methods on short videos where scene change is clear, under which supervised methods could perform better when large-scale labeled training samples are available. However, the collected training samples highly constrain the generability and robustness of those approaches. 

Due to the fact that labeling large-scale long videos is time-consuming and expensive, so the current annotation result can be considered as the average result, which ensures the general quality, while may not preserve the nature that individual annotator may have different preferences, which can be useful as user-study materials.
In future work, we will execute annotations for the same videos by different annotators separately,  for evaluation and verification, which could provide human upper bound and future insights.

\section{Conclusion}
In this paper, we proposed LiveSeg, an unsupervised multimodal framework, focusing on temporal segmentation of long
Livestream video (TSLLV) task, which has not been explored before. 
We collected a large Livestream video dataset named MultiLive, and provided human annotations of 1,000 Livestream videos for evaluation.
By quantitative analysis and human evaluation of our experimental results, we demonstrated that our model is able to generate high-quality temporal segments, which established the basis for Livestream video understanding tasks and can be extended to many real-world applications.

\clearpage
{\small
\bibliographystyle{ieee_fullname}
\bibliography{egbib}
}

\end{document}